%% file: main.tex
\pdfoutput=1

\documentclass[11pt]{article}

\usepackage{acl}

\usepackage{times}
\usepackage{latexsym}
\usepackage{graphicx}
\usepackage{tikz}
\usepackage{xcolor}
\usepackage{todonotes}
\usepackage{xcolor}
\usepackage[normalem]{ulem}
\usepackage[T1]{fontenc}

\usepackage[utf8]{inputenc}

\usepackage{microtype}

\usepackage{inconsolata}

\title{Exploring Large Language Models for Hate Speech Detection in \emph{Rioplatense} Spanish}

\author{
  Juan Manuel Pérez \\ Instituto de Cs. de la Computación \\ CONICET, UBA  \\ \texttt{jmperez at dc.uba.ar} \And
  Paula Miguel \\ Instituto Gino Germani \\ FSOC, UBA \\ CONICET \\ \texttt{paula at sociales.uba.ar} \AND
  Viviana Cotik \\ Instituto de Cs. de la Computación \\ Departamento de Computación, FCEN UBA \\ CONICET, UBA  \\ \texttt{vcotik at dc.uba.ar}
}

\newcommand{\beto}{\textit{BETO}}
\newcommand{\aya}{\textit{Aya}}
\newcommand{\gpt}{\textit{GPT-3.5}}
\newcommand{\mixtral}{\textit{Mixtral}}
\newcommand{\bert}{\textit{BERT}}

\begin{document}
\maketitle
\begin{abstract}
  Hate speech detection deals with many language variants, slang, slurs,
  expression modalities, and cultural nuances.
  This outlines the importance of working with specific corpora, when addressing hate speech within the scope of Natural Language Processing, recently revolutionized by the irruption of Large Language Models. This work presents a brief analysis of the performance of large language models in the detection of Hate Speech for \textit{Rioplatense} Spanish. We performed classification experiments leveraging chain-of-thought reasoning with \textit{ChatGPT 3.5}, \textit{Mixtral}, and \textit{Aya},  comparing their results with those of a state-of-the-art BERT classifier. These experiments outline that, even if large language models show a lower precision compared to the fine-tuned BERT classifier and, in some cases, they find hard-to-get slurs or colloquialisms, they still are sensitive to highly nuanced cases (particularly, homophobic/transphobic hate speech). We make our code and models publicly available for future research.

\end{abstract}

\input{latex/introduction.tex}

\input{latex/related_work.tex}

\input{latex/data.tex}

\input{latex/classification.tex}

\input{latex/results.tex}

\input{latex/conclusions.tex}

\section{Limitations}

One of the main challenges faced in this work is related to the task itself: hate speech detection, which tries to capture a complex social phenomenon. Additionally, it is important to note that the original dataset lacked natural language explanations for the annotations.

Also, the task of regionalism detection could be enhanced, whether by human annotation or by dictionary enrichment, based on human annotations.

Finally, the analysis of LLMs explanations was performed in a very limited way, being their soundness assessed solely by the authors. A deeper analysis of those explanations could be of interest, by including larger samples, more annotators, and the use of other metrics (such as informativeness).

\section{Future work}

As a future work, it would be interesting to develop a multi-variant of Spanish evaluation, as this kind of analytic effort and experiments were out of the scope of this work at this stage. 
In particular, it could be of interest to conduct similar experiments with other Spanish variants, such as Iberian, where there are more available corpora and/or Mexican Spanish, which represents the major variant of spoken Spanish.

It also could be worthwhile to consider regional specificity and/or contextual information, to distinguish text containing challenging elements, such as wordplays, metaphors related to regional knowledge, idiomatic expressions, and instances of irony. Taking that into account, would lead to better identification of regional terms, and future work could be enhanced by exploring in depth different categories of hate speech and the specific use of slang and colloquialisms tied to them. Along with this, future work could focus on improving the prompting engineering to have a better handling of dialectal variants.

\section*{Acknowledgements}

This research has been supported by funds for interdisciplinary projects evaluated and accredited by the Universidad de Buenos Aires, Argentina (PIUBA-2020-3; PIUBA-2022-04-02; PIUBA-2023). The authors would like to thank to the research team members for their collaboration in previous work. They also thank the National Council for Scientific and Technical Research (CONICET) for their support.
The authors are specially grateful to the CCAD, Universidad Nacional de Córdoba,\footnote{\url{https://ccad.unc.edu.ar/}} part of SNCAD–MinCyT, Argentina, for providing access to the computational resources used in this work.

\bibliography{custom}

\appendix

\input{latex/new-appendix}

\label{sec:appendix}
\end{document}

%% file: latex/introduction.tex
\section{Introduction}

In recent years, an increasingly unfolding of violent, discriminatory and hateful speeches can be found on digital platforms, media and networks \cite{berecz2017relevance,woods2019make,hodge2021networks}. While discriminatory discourses emerge in different enunciation areas and modalities, we are particularly interested in the way this kind of problematic speech is spreed along with certain themes in the public arena and the circulation of news \cite{assimakopoulos2017online}. The news published on social networks encourages the debate or discussion of public issues and problems among users. And many times the tweets displaying hate speech are reactions to that news tweets, meaning the context of those messages is extremely important \cite{cotik2020a}. Social media, such as Twitter, offers valuable data access to a relatively natural environment for the study of hate speech, being particularly interesting the activation of hate speech regarding public topics, for example topics triggered by news that are subject of discussion \cite{zannettou2020measuring, erjavec2012you}.

Hate speech detection, from the Natural Language Processing (NLP) perspective, has to deal with languages crossed by variants, slang, slurs, and other specific modalities found \cite{nunberg2018social, diaz2020slur}. Being aware of cultural nuances and specific contexts of use is crucial to address this task and shows the relevance of developing dialectal corpora and analysis that allows to automatically detect specific hateful expressions in different lexical contexts. 

Regarding this issue, a particular interest arises in the performance of large language models (LLMs) by analyzing hate speech in relation to local expression nuances. LLMs have shown to be effective in a wide range of NLP tasks \cite{brown2020language,wei2021finetuned,ouyang2022training}. 
Given that GPT-3.5 (also known as ChatGPT) is one of the most popular and emerging large language models \cite{wu2023brief, deng2022benefits}, we have the question of how well it can detect hateful messages in a specific dialectal variant of Spanish, focusing on the \textit{Rioplatense} variety.


With over 450 million native speakers, primarily in Latin America, Spain, and also parts of the US  \cite{ethnologue}, Spanish includes many varieties and dialects. Each one represents a common cultural background and semantic field, expressing different uses for some words or, contrary wise, the use of specific words or phrases addressing the same purpose. Among them, \textit{Rioplatense} Spanish, mainly spoken both in Argentina and Uruguay, is thought to be spoken by more than one-tenth of Spanish native speakers \cite{lipski2012geographical, coloma2018argentine}.

One of the most interesting features of \textit{Rioplatense} Spanish is that includes ruralisms, indigenisms, argot and slang, especially \textit{lunfardo}, among other vocabularies. While almost all languages have repertoires of expression outside of general use, the case of \textit{lunfardo} constitutes a linguistic phenomenon in which words and expressions of very diverse origin converge (Italian, popular Spanish, French, Portuguese, Guaraní, Quechua, among others), as result of the migratory processes in Argentina, with its epicenter in Buenos Aires, especially during the 19th century and the first half of the 20th century \cite{de2003diccionario, conde2013lunfardo}. \textit{Lunfardo} shows as an integrated lexical repertoire, which has around 6,000 voices, of which only about 300 are recognized by the Dictionary of the Royal Spanish Academy,\footnote{The Royal Spanish Academy (RAE) is a cultural institution dedicated to linguistic regularization in the Spanish-speaking world.} such as the term "\textit{laburar}", from Italian "\textit{lavorare}", standing as the verb "to work" or "\textit{trabajar}" in Spanish. Many words have other meanings, for example "\textit{trola}" means a lie o deceit in Iberian Spanish, but in \textit{Rioplatense} Spanish it refers to "\textit{whore}", and it also has a masculine form "\textit{trolo}" that means "\textit{faggot}". These terms are commonly used to derogatorily address both women and gays. These examples point out that the understanding of many expressions constitutes a challenge even to Spanish native speakers unfamiliar with the \textit{Rioplatense} variant. Would that be the case of LLMs?

Following that question, this paper develops an exploratory approach to the effectiveness of LLMs addressing the task of detecting specific texts and tagging corpora linked to hate speech. We take as a benchmark a fine-tuned BERT classifier trained with a corpus of news articles and Twitter's comments in \textit{Rioplatense} Spanish, annotated to detect hate speech \cite{perez2023assessing}. We performed classification experiments leveraging the chain-of-thought (CoT) reasoning within the LLMs \textit{ChatGPT}, \textit{Mixtral} and \textit{Aya}, and compare their results against a fine-tuned BERT classifier. Our experiments point out that LLMs show a lower precision compared to the fine-tuned BERT classifier, but a higher recall for highly nuanced cases (particularly, homophobic/transphobic hate speech). However, explanations given by ChatGPT are ---while not equal to human annotators--- convincing in most cases.

The results of this work, focused in \textit{Rioplatense} Spanish, contribute to understanding possible forms of bias and to evaluate the specificity of \textit{Rioplatense} variant that is not part of the typical corpus used for training LLMs. This highlights its unique characteristics and the importance of developing and working with dialect-specific corpora for the automated detection of hate speech. Focusing exclusively on \textit{Rioplatense} Spanish is a first step in addressing cultural nuances and specific local expressions inherent in this dialect, particularly in relation to the performance of LLMs. Future work could be pursued on this basis, aiming to compare Rioplatense Spanish with other variants, such as Mexican or Iberian Spanish.


%% file: latex/related_work.tex
\section{Related work}

There is a plethora of resources for automatic detection of hate speech in English. Nevertheless, when it comes to Spanish, despite being one of the main languages in terms of the number of native speakers worldwide \cite{ethnologue}, corpora are scarce, with only a few datasets publicly available. Some of the main references of published and available resources in Spanish are found in datasets like \textit{IberEval}, which presents a \textasciitilde 4k Twitter dataset for the Automatic Misogyny Identification (AMI) \cite{fersini2018overview}; those released by the MEX-A3T task, that included a dataset of \~11k Mexican Spanish tweets annotated for aggressiveness \cite{alvarez2018overview}; and the dataset launched in the context of the HatEval challenge, a \textasciitilde 6.6k tweets dataset annotated for misogyny and xenophobia \cite{basile2019semeval}. To the best of our knowledge, only one dataset annotated for hate speech detection is available in \textit{Rioplatense} \cite{perez2023assessing}. This highlights the positive impact of contributions tending to build resources in Spanish and its variants.

Regarding the automatic detection and treatment of hate speech, a broad amount of literature has been published. We refer the readers to \citet{poletto2021resources,schmidt2017survey,fortuna2018survey} for extensive reviews of work in the field of NLP. In this section, we focus on the most recent work on hate speech detection, explanation, and treatment using LLMs.

Upon the blazing development of LLMs \cite{brown2020language,wei2021finetuned,ouyang2022training}, some studies have been conducted to evaluate their performance in hate speech detection, explanation and treatment. \citet{sap2020social} used GPT-2 to detect and generate hate speech explanations. \citet{plazadelarco2023respectful} assessed the performance of several language models (such as the instruction-finetuned \textit{mT0} \cite{muennighoff-etal-2023-crosslingual} and \textit{FLAN-T5} \cite{chung2022scaling} in zero-shot setting over several hate speech and toxicity datasets. \citet{wang2023evaluating,huang2023chatgpt} evaluated the performance of GPT-3/GPT-3.5 to detect and explain hate speech messages, finding that LLM-generated explanations are equally good (and even preferred to) human-written explanations. Some of these explanations are inducted by chain-of-thought reasoning \cite{wei2022chain}, also known as the ``let's think step by step'' technique. \citet{oliveira2023good} tested ChatGPT for hate speech detection in Portuguese, particularly in its Brazilian dialect, achieving almost state-of-the-art results in a zero-shot setting. \citet{ccam2023evaluation} performed experiments for Turkish, with similar results.


%% file: latex/data.tex
\section{Data}

Our experiments are based on an anonymized dataset available in \textit{Rioplatense} Spanish, consisting of Twitter replies to posts from Argentinean news outlets; with models, partitions and class distributions arealso available \cite{perez2023assessing}. In this dataset, comments to news posted by regional users were annotated for the presence of hate speech and categorized into one or more of eight possible types: politics, misogyny, homophobia/transphobia, racism/xenophobia, class hatred, appearance, hate against criminals and disabled people. All annotated instances have a context (the tweet posted by the news outlet, plus the news title and whole text content of the news) and the text being analyzed and annotated for hate speech (each Twitter user's comment). This is important as contextual information situates the comment and that it has been shown to be relevant to detect hate speech \cite{sheth2021defining, xenos-2021-context}.  Along with the annotated dataset, another unannotated dataset suitable for unsupervised training is provided. The unsupervised corpus used to continue pre-training includes around 5,000,000 \textit{Rioplatense} tweets comments, with a contextual information of 288,000 news tweets and full articles.

In this work, we address the presence of hate speech linked to four possible types: misogyny, homophobia/transphobia, racism/xenophobia, and class hatred/classism, according to the attacked characteristics, from now on dubbed WOMEN, LGBTI, RACISM, and CLASS (see Appendix \ref{subsec:classes} for a broader definition). The selection of those categories was made based on the prevalence of each category among hateful speech and their societal impact. Also, these topics are widely covered and considered in the available literature, meaning the results could be a useful contribution to standard ground and state of the art  \cite{paz2020hate, tontodimamma2021thirty}.

We used the same train and development samples as \citet{perez2023assessing} ($36,420$ and $9120$ examples respectively), but, for budget reasons, we subsampled the test set to $5670$ examples ($50\%$ of the original test set). Among these, $479$ comments contain hate speech, distributed as shown in Table \ref{tab:count}. Some comments express attacks to more than one category. In those cases, we find two relevant combinations: RACISM associated with CLASS (21 cases), followed by the association of WOMEN and LGBTI (10 cases). Only one comment targeted 3 categories at the same time. Table \ref{tab:dataset_examples} shows some examples of the dataset.

    \begin{table}[b]
        \centering
        \begin{tabular}{lcl}
 Category&Number & Percentage\\\hline
         RACISM& 230 &4.06 \%\\
             WOMEN& 131 &2.31 \%\\
             LGBTI& 88 &1.55 \%\\
             CLASS& 76 &1.34 \%\\\hline
 TOTAL& 479&8.45 \%\\
        \end{tabular}
        \caption{Number and percentage of messages containing hate speech in each category.}
        \label{tab:count}
    \end{table}

\begin{table*}[t]
    \centering
    \footnotesize
    \begin{tabular}{p{0.1\textwidth} p{0.4\textwidth} p{0.3\textwidth}}
        Category              & Context                                                                                                  & Comment                          \\
        \hline
        WOMEN                 & Mia Khalifa: acted in porn videos for a few months, became world famous and now fights to erase her past & HAHAHA KEEP SUCKING....          \\
        \rule{0pt}{3ex}LGBTI  & The story of the Colombian trans model kissing the belly of her eight-month pregnant husband             & A male kissing another male      \\

        \rule{0pt}{3ex}RACISM & Yanzhong Huang: ``It is quite likely that a Covid-21 is already brewing''                                & Urgent bombs to that damned race \\
        \rule{0pt}{3ex} CLASS & Social movements cut off 9 de Julio Av.: they demand a minimum wage of \$45,000                          & get to work, mfs                 \\
        \hline
    \end{tabular}
    \caption{Hateful examples from the analyzed dataset.}
    \label{tab:dataset_examples}
\end{table*}




In order to assess whether the large language models were able to capture the meaning of regional terms and expressions or not, we marked the test dataset for the presence of regionalisms. We referred to regionalisms as idiomatic phrases, words exclusively used in \textit{Rioplatense} Spanish, or with a meaning that is differentially used in Argentina or only in \textit{Rioplatense}-speaking countries (eg. "pelotudo" for "asshole").
As aforementioned, there is a vacancy area regarding \textit{Rioplatense}:  resources are scarce and, when available, they are not the best to work with. Therefore, we generated a very basic and not exhaustive list of regional terms, scraped from an crowd-sourced dictionary of regionalisms in \textit{Rioplatense} Spanish.\footnote{ \url{https://www.diccionarioargentino.com/}}


%% file: latex/classification.tex
\section{Classification experiments} 


\label{sec:classification}
%
%
%

We compared two kinds of classification algorithms:

\begin{itemize}
    \item Pre-trained language models based on \bert{}: fine-tuned on supervised data from the corpus.
    \item Large Language Models (LLMs) using few-shot learning and chain-of-thought reasoning (CoT) \cite{wei2022chain}.
\end{itemize}

For the first group of classifiers, we tested two pre-trained models in Spanish: \beto{} \cite{canete2020spanish}, and RoBERTuito \cite{perez2022robertuito}. For each model, we performed a small hyperparameter search following the guidelines of \citet{tuningplaybookgithub}, searching for the best-performing values for the number of epochs, the learning rate and warm-up ratio. To track our experiments, we used the \textit{wandb} library \cite{wandb}. For each of the pre-trained models, we previously fine-tuned them on the unsupervised corpus provided along with the used dataset \cite{perez2023assessing}, as it has been shown to improve the performance in domain-specific tasks \cite{gururangan2020don}.


The fine-tuning process of the supervised models followed standard practices in fine-tuning BERT-like models.  The classifiers were trained with Adam \citep{kingma2014adfam} as the optimizer with weight decay, and a triangular learning rate schedule. 

\begin{table}[b]
    \centering
    \footnotesize
    \begin{tabular}{cc}

        \textbf{Hyperparameter} & \textbf{Values}                          \\
        \hline
        Epochs                  & 3, 4, 5                                  \\
        Batch Size              & 32                                       \\
        Learning Rate           & 2e-5, 3e-5, 5e-5, 6e-5, 7e-5, 8e-5, 1e-4 \\
        Weight Decay            & 0.1                                      \\
        Warmup Ratio            & 0.06, 0.08, 0.10                         \\
        \hline
    \end{tabular}
    \caption{Hyperparameter search space considered for each model.}
    \label{tab:hyperparameters}
\end{table}




Table \ref{tab:hyperparameters} outlines the spectrum of values applied to each hyperparameter. For every model, task, and language, we conducted between 30 and 60 runs, choosing the optimal model based on the Macro F1 score from the validation set. 
We adopted a batch size of 32, tailored to accommodate our GPU memory limitations (either a GTX 1080Ti or Tesla T4, with memory ranging from 11 to 14GB). The best hyperparameters found in the tuning process were roughly the same for all the models: $0.1$ for warm-up ratio, $3$ and $4$ (only RoBERTuito on its non-finetuned version) the number of training epochs, and finally $6e^{-5}$ for the learning rate.

\subsection{Prompting strategies}
Regarding the large language models, we selected three models that show good performance in Spanish to run the experiments and prompts:

\begin{itemize}
    \item \gpt{} turbo-0125 \cite{ouyang2022training}: a closed-source large language model provided by OpenAI, that has an outstanding performance in several tasks.
    \item \mixtral{} \cite{jiang2024mixtral}: a mixture-of-experts open-source language model pre-trained in English, French, Italian, German and Spanish.
    \item \aya{} \cite{ustun2024aya}: a massively-multilingual sequence-to-sequence language model, that follows the architecture of \textit{T5} \cite{raffel2020exploring}, pre-trained in 101 languages.
\end{itemize}

Mixtral and Aya were run in two NVIDIA A30, using the Transformers library. The same prompt was used for the three LLMs. 

To build the prompt, we resorted to few-shot or in-context learning. Early experiments using zero-shot, one-shot settings and few-shot learning were conducted. Then, different prompts were evaluated in the development split until no improvement in performance was observed.

That prompt engineering process led to the following prompt:\footnote{Originally in \textit{Rioplatense} Spanish (see in Appendix), translated to English for the purpose of this paper.}
\begin{quote}

    Determine if the following text, corresponding to a tweet, presented with a context,  contains hate speech. We understand that there is hate speech if it has statements of an intense and irrational nature of rejection, enmity, and abhorrence against an individual or against a group, being the targets of these expressions for possessing a protected characteristic. The protected characteristics we consider are:

    \begin{itemize}
        \item WOMEN: refers to women or the feminist movement
        \item LGBTI: refers to gays, lesbians, transgender individuals, and other gender identities
        \item RACISM: refers to immigrants, xenophobia, or against indigenous peoples
        \item CLASS: refers to low-income people or class-related issues
    \end{itemize}

    The tweets are written in Rioplatense Spanish, and within the cultural context of Argentina. Respond with one or more of the characteristics separated by commas, or "nothing" if there is no hate speech. Think and justify the response step by step before answering.
\end{quote}

We leveraged chain-of-thought reasoning \cite{wei2022chain} to both enhance the model's performance and to provide an explanation for the prediction. The model was prompted with a total of 12 examples of hate speech considering the different characteristics. The examples were selected from the training set, and consisted of three lines, such as this:

\begin{quote}
    \textbf{context:} Wuhan celebrates the end of the coronavirus quarantine with a message for the rest of the world: ``Learn from our mistakes''

    \textbf{text:} Motherfuckers! I wish you all chinese people die

    \textbf{output:} The text wishes that Chinese people would die, blaming them for the COVID-19 pandemic. answer is ``racism''.

\end{quote}

The output consisted of a natural language explanation. 


The 12 examples considered the different characteristics and target labels and were balanced by 
their labels. They 
also included 
“nothing” examples, that pointed out cases where there was no hate speech towards the targeted categories (women, LGBTI, racism or class). The selection of examples provided in the prompt
had the following distribution of categories: 2 examples for racism, 2 for LGBTI, 2 for women and 2 for "nothing";  1 example for classism; and 1 multi-class example for racism and class. The full list of examples and the original prompt in Spanish can be found in Appendix  \ref{subsec:prompt}.

\subsection{Evaluation}

To evaluate the performance of the classifiers, we assessed the precision, recall, and F1-score for each class, in a multi-label classification schema. We get bootstrap 95\%-CI intervals using the \textit{confidence-intervals} library \cite{ferrer2023confidence}. We also evaluated a subset of the dataset, that specifically contains regional terms.


%% file: latex/results.tex
\section{Results}

\begin{table}
    \centering
    \tiny
    \begin{tabular}{ccccc c}
                        & WOMEN & RACISM & LGBTI & CLASS & Macro \\
        \hline
        BETO            & 0.366 & 0.698  & 0.414 & 0.473 & 0.570 \\
        RoBERTuito      & 0.414 & 0.675  & 0.435 & 0.451 & 0.582 \\
        BETO (FT)       & 0.422 & 0.736  & 0.468 & 0.511 & 0.614 \\
        RoBERTuito (FT) & 0.405 & 0.694  & 0.474 & 0.471 & 0.598 \\
        \hline
    \end{tabular}
    \caption{F1 scores for the considered BERT classifiers. FT stands for fine-tuned, marking those pre-trained models that were further fine-tuned in the unsupervised corpus.}
    \label{tab:bert_full_results}
\end{table}

Table \ref{tab:bert_full_results} shows the F1 scores for all the considered BERT classifiers. As mentioned in Section \ref{sec:classification}, we considered BETO \cite{canete2020spanish} and RoBERTuito \cite{perez2022robertuito}, and also their fine-tuned versions in the unsupervised corpus provided by the considered dataset \cite{perez2023assessing}. We can observe that, although RoBERTuito performs better in the non-finetuned version, BETO achieves the best results after the pre-training process.

\begin{table}[t]
    \centering
    \begin{tabular}{llll}
        \hline
                   & F1             & Precision      & Recall         \\
        Model      &                &                &                \\
        \hline
        \aya{}     & $21.2 \pm 0.8$ & $11.9 \pm 0.5$ & $93.0 \pm 1.2$ \\
        \gpt{}     & $47.8 \pm 1.8$ & $39.2 \pm 1.8$ & $61.2 \pm 2.2$ \\
        \mixtral{} & $38.6 \pm 1.3$ & $25.1 \pm 1.0$ & $83.8 \pm 1.7$ \\
        \beto{}    & $63.5 \pm 1.8$ & $72.9 \pm 2.4$ & $56.3 \pm 2.1$ \\
        \hline
    \end{tabular}

    \caption{Classification results for LLMs and fine-tuned BETO, expressed as macro averages of F1, Precision and Recall for all the considered labels.}
    \label{tab:results}
\end{table}

Table \ref{tab:results} shows the results for the multi-label classification task, represented as the macro averages of F1, Precision and Recall. We report only the results of the best BERT classifier, the fine-tuned version of BETO. While the BETO classifier outperforms their counterparts in terms of precision and F1, the LLMs  have higher recall. As Aya model performed poorly, qualitative analysis is focused on the LLMs GPT-3.5 and Mixtral.

\begin{figure}[t]
    \centering
    \hspace{-0.4cm}\includegraphics[width=0.5\textwidth]{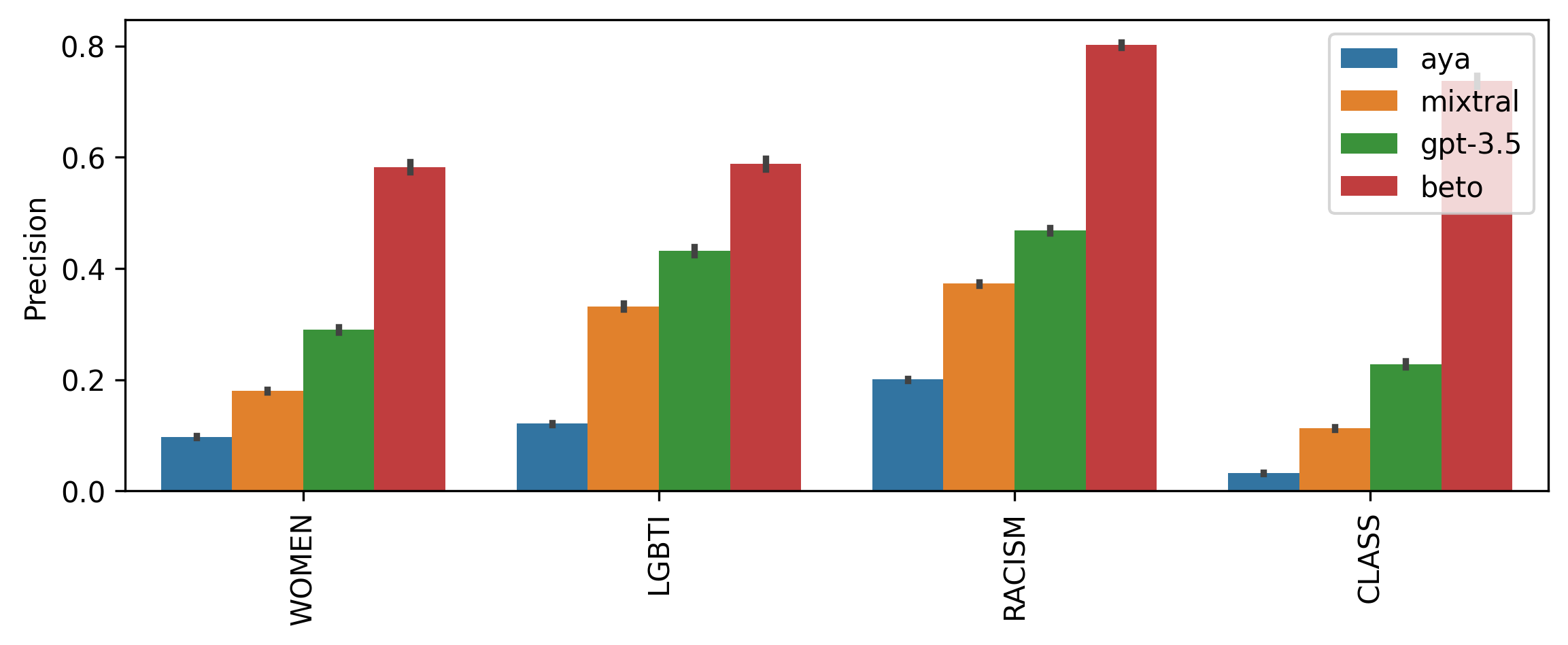}

    \hspace{-0.4cm}\includegraphics[width=0.5\textwidth]{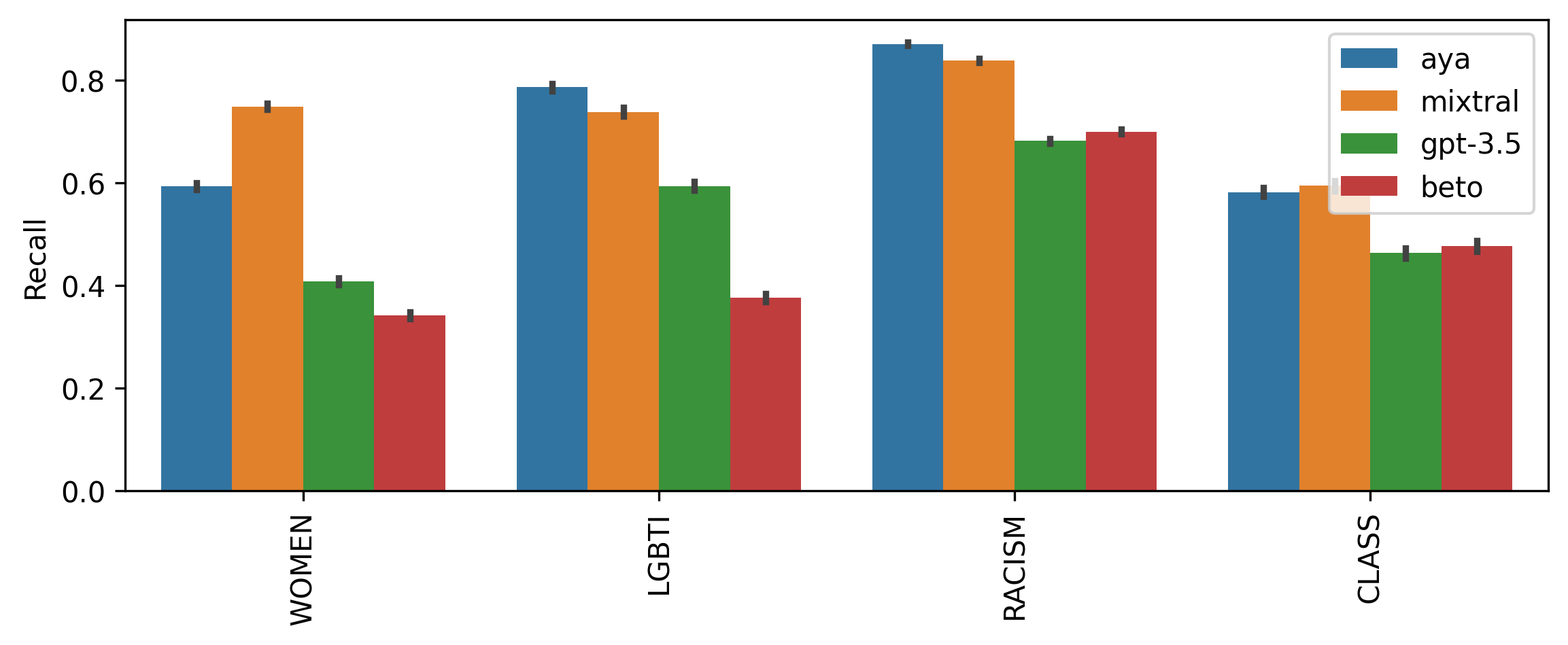}

    \hspace{-0.4cm}\includegraphics[width=0.5\textwidth]{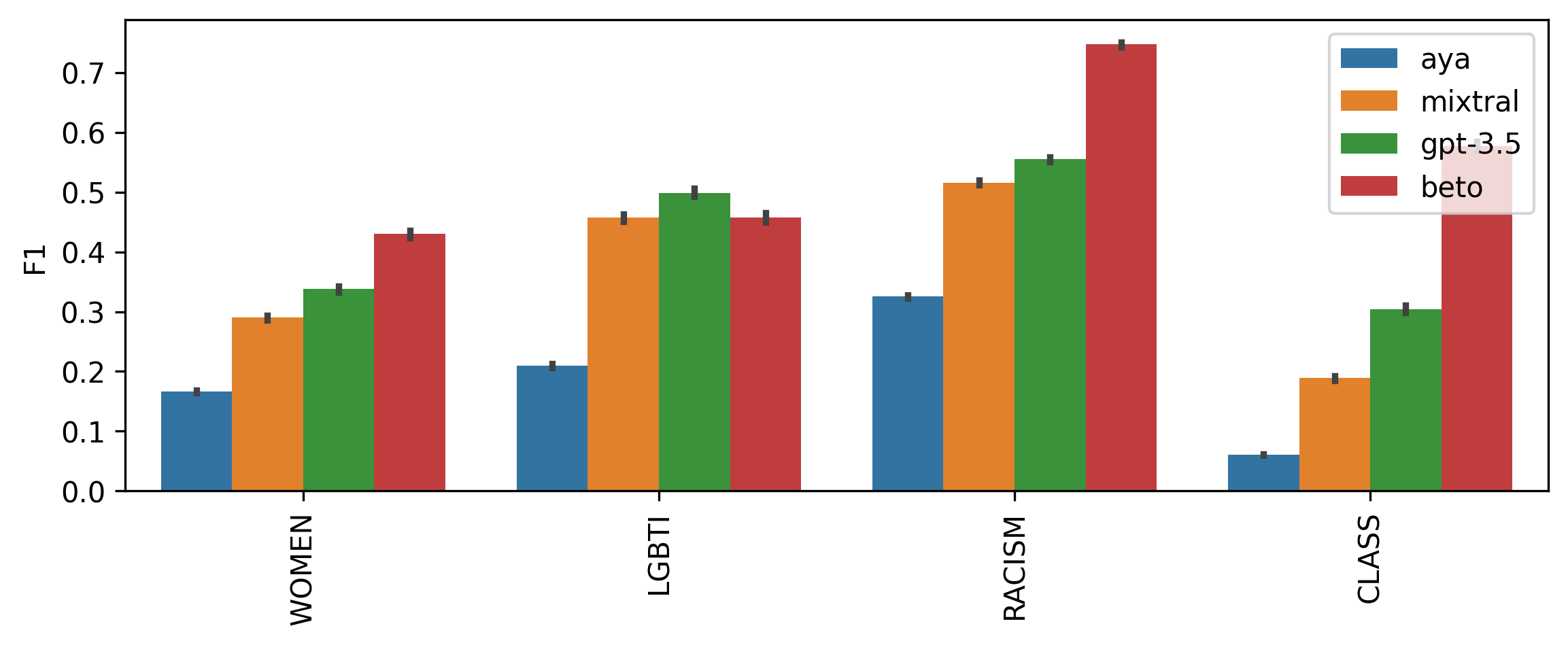}
    \caption{Precision, recall and F1 of the classifiers: ChatGPT 3.5, Aya, Mixtral and the fine-tuned \beto{} classifier.}
    \label{fig:results}
\end{figure}

A closer inspection of each of the considered characteristics is presented in the multi-label classification results, shown in Figure \ref{fig:results}. It is shown that \mixtral{} obtains a better recall for all of the characteristics but at the cost of low precision, while \gpt{} has a better trade-off between them.  The case of the LGBTI characteristic, is particularly interesting given that is the case where \gpt{} outperforms the fine-tuned classifier ($F1 = 49.9 \pm 4.4$ for \gpt{} and $F1 = 45.7 \pm 5.1$ for \beto{}, Mann-Whitney U=$16386.5, p \leq 1 * 10^{-6}$). This is particularly relevant as this characteristic is difficult to detect, as it often involves culturally complex language, irony and metaphors and where \bert{}-based classifiers show a significant gap compared to humans \cite{yigezu2023transformer, perez2023assessing}.

Focusing in those messages that contain regionalisms, no conclusive differences are detected. As seen in \ref{tab:regionalism}, in general, the performance of all the classifiers follows similar patterns: they stay equal or get better when regionalisms are present in the text.
This might be due to the fact that regionalisms are likely to be matched with slang slurs which both the \beto{} and the LLMs (particularly, \gpt{}) leverage for this detection task. Indeed, hate speech messages represented 9\% of the original dataset, and this rises to a 16\% of hateful messages within those that have regionalisms.

Following up, for example, in CLASS category, words such as \textit{planero, villero}, constituted part of the repertoire of classist denigrational speech; and regionalisms in form of slurs against WOMEN (\textit{trola, yegua, abortera}) are also very common. These two classes are, in fact, those with more hateful comments containing regionalisms: 71\% of CLASS hateful messages contains regionalisms and the same account for 62\% of those against women. These slur, regional expressions may make the task of flagging hateful speech easier compared to comments lacking it.

\begin{table}
    \centering
    \scriptsize
    \begin{tabular}{c cc cc cc}
        Model                    &
        \multicolumn{2}{c}{BETO} & \multicolumn{2}{c}{GPT-3.5} & \multicolumn{2}{c}{Mixtral}                                     \\\hline
        Regionalism              & reg.                        & wo. reg.                    & reg. & wo. reg. & reg. & wo. reg. \\\hline
        CLASS                    & 0.67                        & 0.30                        & 0.46 & 0.11     & 0.28 & 0.09     \\
        LGBTI                    & 0.47                        & 0.44                        & 0.50 & 0.48     & 0.44 & 0.47     \\
        RACISM                   & 0.70                        & 0.76                        & 0.60 & 0.54     & 0.56 & 0.50     \\
        WOMEN                    & 0.51                        & 0.27                        & 0.42 & 0.25     & 0.39 & 0.20     \\
        \hline
        Macro                    & 0.60                        & 0.48                        & 0.50 & 0.37     & 0.42 & 0.32     \\
        \hline
    \end{tabular}
    \caption{
        F1 by category for messages containing regionalism (reg.) and those not containing it (wo. reg.).}
    \label{tab:regionalism}
\end{table}

\begin{table*}[t]
    \centering
    \scriptsize
    \begin{tabular}{p{0.2\textwidth} p{0.2\textwidth} p{0.25\textwidth} p{0.25\textwidth}}

        \textbf{News Context}                                                                                                                                                                                                                                             & \textbf{Reaction Tweet}                                                                                                                                                                                                                                                            & \textbf{GPT-3.5 output}                                                                                                                                                                                                                        & \textbf{Mixtral output}                                                                                                                                                                              \\
        \hline
        \multicolumn{4}{c}{\beto{} false negatives correctly classified by LLMs}                                                                                                                                                                                                                                                                                                                                                                                                                                                                                                                                                                                                                                                                                                                                                                                                                                                                                                                                       \\

        \hline
        After a barrage of more than 20 shots, a trans woman was murdered in Rosario                                                                                                                                                                                      & The prostate should not appear in the autopsy by law!                                                                                                                                                                                                                              & The message refers to the gender identity of the trans woman murdered in a discriminatory manner, using an emoji to mock the situation. The final answer is "LGBTI".                                                                           & The text refers to the victim as "a man" and denies her gender identity. The final answer is "LGBTI".                                                                                                \\

        \hline

        ¿More Rial found love in a personal trainer? The mediatic shared a live broadcast with a handsome personal trainer where they looked very intimate                                                                                                                & That one sees a \textit{pingo} (Arg. for "horse", but also "dick") and jumps on.                                                                                                                                                                                                   & The text refers to More Rial's sexuality in a derogatory way, using the term "pingo" in a pejorative way to refer to a penis. The final answer is "woman".                                                                                     & The text refers to More Rial as "that one" and accuses her of getting on a man because of her physical attractiveness. This is sexist and derogatory. The final answer is "woman".
        \\

        \hline
        Mara Gómez: the story of the first trans soccer player in the Argentine championship                                                                                                                                                                              & \textit{Ponga huevos, Mara ponga huevos...} (Eggs on [winning], Mara, Eggs on! Here "eggs" stands for testicles)                                                                                                                                                                   & The text cheers Mara Gómez as a soccer player, but she uses the expression "\textit{ponga huevos}" in a discriminatory way, with the aim of questioning her masculinity and her gender identity as a trans woman. The final answer is "LGBTI". & The text alludes to Mara Gómez's lack of "huevos", a term used to refer to the lack of courage or bravery. This is offensive and derogatory. The final answer is "woman, LGBTI."                     \\

        \hline


        \multicolumn{4}{c}{Incorrectly classified or  incorrect explanations by LLMs}                                                                                                                                                                                                                                                                                                                                                                                                                                                                                                                                                                                                                                                                                                                                                                                                                                                                                                                                  \\
        \hline

        British arrived in Peru for 10 days and stayed for love: she lives without safe water and among insects

                                                                                                                                                                                                                                                                          & The Peruvian's "\textit{gaver}". Mother. ("gaver" stands for "verga", meaning dick. It is a sexist comment, not racist)
                                                                                                                                                                                                                                                                          & The text contains a derogatory comment towards Peruvians, using a vulgar and discriminatory term. The final answer is "racism."
                                                                                                                                                                                                                                                                          &
        The text refers to Peruvians as "gaver", a derogatory term. The final answer is "racism."
        \\

        Mara Gómez: the story of the
        first trans soccer player in the Ar-
        gentine championship                                                                                                                                                                                                                                              & It comes to bring what the team lacks: goals and a lot of "eggs" (in the sense of courage or bravery, refering to testicles. The comment should be tagged as LGBTI.)                                                                                                               & The text praises Mara Gómez's soccer skills and mentions that she plans to contribute goals and effort to the team. It does not contain hate speech or mention any protected characteristics. The final answer is "nothing."                   & The text alludes to Mara Gómez's ability as a footballer, without making any allusion to a protected characteristic. The final answer is 'nothing'.
        \\
        \\
        \hline

        Tension in Villa Mascardi: Mapuches resist the eviction of a usurped property.
                                                                                                                                                                                                                                                                          & How strange, [Joe] Lewis and the millionaire squatters are not going to be evicted.
                                                                                                                                                                                                                                                                          & The text mentions the Mapuches in a derogatory manner and suggests preferential treatment towards non-indigenous occupants. Although protected characteristics are not directly mentioned, the comment denotes prejudice towards the Mapuche people. The final answer is "racism."

                                                                                                                                                                                                                                                                          & The text compares the Mapuches with the "millionaire squatters", suggesting that they are equal. This shows contempt for indigenous peoples and their fight for their lands. The final answer is "racism."
        \\

        \hline
        \multicolumn{4}{c}{Regional slangs and some cultural nuances detected by LLMs}                                                                                                                                                                                                                                                                                                                                                                                                                                                                                                                                                                                                                                                                                                                                                                                                                                                                                                                                 \\
        \hline
        Hebe de Bonafini\footnote{One of the founders of the Association of the Mothers of Plaza de Mayo, mothers whose children disappeared during the 1976-1983 dictatorship.} criticized supermarkets and called for “replacing meat with rice, chickpeas and lentils” & @usuario Old "chupa pi." ("pi." stands for "pija" ("cock"), meaning "cocksucker"), you like "morcilla" (literaly "blood sausage", but also "dick").                                                                                                                                & The text contains an insult towards Hebe de Bonafini, calling her an "vieja chupa pi." (old cocksucker), but also makes derogatory reference to her diet, using a sexist and vulgar comment. The final answer is "woman."                      & The text insults Hebe de Bonafini by calling her a "vieja chupa pi." (old cocksucker), a derogatory expression. Additionally, she makes fun of her dietary preferences. The final answer is "woman." \\

        \hline

        China: another virus identified “with the potential to become pandemic” \#NewVirus \#China                                                                                                                                                                        & First they \textit{morfan} (Arg. lunf. "eat") bats, now they must have \textit{garchado} (Arg. "fucked") a pig.                                                                                                                                                                    & The text contains a derogatory comment towards Chinese, alluding to their eating habits and suggesting a repulsive sexual intercourse with a pig. The final answer is "racism".                                                                & The text shows contempt for Chinese culture and its relation to animals. The final answer is "racism."                                                                                               \\
        \hline
    \end{tabular}

    \caption{Examples of LLMs results}
    \label{tab:examples}

\end{table*}


In the case of LGBTI hateful instances, when hate speech detection task faces subtle, non-slur, slang related language attacking transgender people, most explanations generated by LLMs were correct, getting to detect difficult-to-tag messages that were overlooked by the \beto{} classifier. Table \ref{tab:examples} shows some examples of those cases.\footnote{The analysis is shown for GPT 3.5 and Mixtral (with the benchmark of fine-tuned Beto), as Aya underperformed at this task.}

These results suggest that despite possible preconceptions about the lack of cultural grounding of LLMs, they are somehow able to grasp cultural nuances from \textit{Rioplatense} Spanish and Argentinian culture, even compared to fine-tuned BERT modeled to address that specific dialect. Some of the examples in Table \ref{tab:examples} show that LLMs, sometimes are able to detect, explain, and provide information about regional slang not detected by the fine-tuned \beto{} classifier.

However, LLMs show a higher rate of false positives than the fine-tuned \beto{} classifier, when it comes to the reference of majority-vote labels in the dataset.\footnote{In the original dataset, each comment was annotated by three annotators. Therefore, it was used a majority-vote label.} This might indicate, first, that these models are more sensitive to the presence of hate speech and toxicity (probably due to preference tuning or some other safety mechanisms
)
and second, that the comparison against one single binary label might not be the best way to evaluate these models. Evaluating some of their explanations with other metrics, such as human evaluation of soundness, informativeness, among others \cite{wang2023evaluating}, or also by using a perspectivist framework taking into account the disagreement of the annotations \cite{sachdeva2022measuring, basile2021need} may provide a better comparison between these models.

%% file: latex/conclusions.tex
\section{Conclusions}
The culture and communication of Latin America is diverse and full of different expressions, idioms, slang, specific uses, and adaptations of, among others, the Spanish language, which embodies subtle differences that are often context-dependent and cannot be captured outside of their context of use. We believe this work is a first step toward filling a gap in resources regarding {\textit{Rioplatense}, a particular variety of Spanish, and aims to contribute to the understanding of how large language models perform in these specific,  "under-resourced" language varieties and their cultural contexts, which is particularly important when addressing the phenomenon of hate speech. 

This study shows the effectiveness of \textit{ChatGPT} and \textit{Mixtral} classifiers in the task of hate speech detection, specifically in the context of cultural and linguistic complexities present in \textit{Rioplatense} Spanish tweets. The findings highlight that while \textit{Mixtral} excels in recall across various characteristics, it does so at the expense of precision. Conversely,  \textit{ChatGPT}  offers a more balanced performance, particularly in detecting the LGBTI characteristic, where it outperforms traditional fine-tuned classifiers, such as BETO.  Comparing LLMs with a state-of-the-art fine-tuned BETO classifier, \textit{ChatGPT} and \textit{Mixtral} showed a lower precision but a higher recall in some categories, particularly in difficult cases that the supervised classifier was not able to detect. 
These results underline the potential for LLMs to capture cultural nuances and regional expressions effectively and to interpret culturally sensitive language rich with ironies and metaphors. The results of \textit{ChatGPT}  to identify hate speech laden with regionalisms show ability in processing colloquial language. Notably, the prevalence of regional slurs in classist and misogynistic discourse appears to indicate the importance of incorporating regional context to improve hate speech detection. However, the observed higher rate of false positives in LLM classifications relative to BETO emphasizes the need for careful evaluation metrics when assessing the efficacy of these models.

While LLMs have proven to be a powerful tool for hate speech detection, supervised classifiers still outperform them in the general case and are more suitable for detecting hate speech at large scale. Regarding cultural and linguistic nuances, we found that LLMs were able to detect some of them, but not all, missing some slurs, expressions and insults typical of the \textit{Rioplatense} dialect.

The insights from this study emphasize the importance of building models that are sensitive to cultural and linguistic diversity, while also highlighting the value of producing corpora on specific topics and linguistic variants that serve as benchmarks.
In that sense, this paper contributes to variant-focused research in NLP,  being in this case \textit{Rioplatense} Spanish, underlining the importance of a comprehensive understanding of variants, its cultural nuances and language uses as a vital step in addressing hate speech in a linguistically diverse context. 





%% file: latex/new-appendix.tex
\section{Appendix}
\label{sec:appendix}

In this appendix, we present the definition of each targeted class of hate speech and we describe details of the original prompt and instruction provided to the LLMs 

\subsection{Classes Definition}

\label{subsec:classes}

Definition of each class addressed in hate speech detection according to the  \textit{Plan Nacional contra la Discriminación} (Argentinean National Plan Against Discrimination) guidelines \cite{2006hacia}.

\subsubsection{RACISM}
Racism can be defined as a set of beliefs, attitudes and practices that discriminate against individuals or groups based on their race, ethnicity, or physical characteristics, denying them equal rights and dignity. This discrimination can be both individual and institutional and can occur in various ways: Individual racism; Institutional racism; Structural racism; Cultural racism; Xenophobia (rejection or fear of people from other countries or cultures); and also includes Antisemitism, Arabophobia, Islamophobia and Afrophobia.
\subsubsection{WOMEN}
Discrimination against women is a social phenomenon that manifests itself as a set of actions and beliefs that devalue women compared to men, through attitudes, practices and structures that perpetuate gender inequality and limit women's access to resources, rights and opportunities in various areas of life. Discrimination against women can be expressed in different areas, such as Economic Inequality; Gender Violence; Access to Education; Political Representation; Cultural Norms and Stereotypes; among others. Added to this is intersectional discrimination (women who belong to minority groups may experience additional forms of discrimination, which further complicates their situation).
\subsubsection{LGBTI}
Gender-based discrimination refers to attitudes, behaviors, and policies that marginalize and devalue people based on their sexual orientation or gender identity. This form of discrimination manifests itself in various dimensions and contexts, affecting the daily lives and fundamental rights of LGBTQ+ people. Gender-based discrimination can be expressed as violence and harassment; legal inequality; social stigmatization; unequal access to services; non-positive media representation; among others. Likewise, gender-based discrimination can intersect with other forms of discrimination, such as racism or poverty, exacerbating the difficulties faced by individuals belonging to multiple marginalized groups.
\subsubsection{CLASS}
Class-based discrimination refers to attitudes, practices, and social structures that marginalize people based on their socioeconomic status/poverty. This form of discrimination manifests itself in various dimensions, affecting access to resources, rights, and opportunities, and perpetuating social exclusion. Expressing itself, for example, as Limited Access to Resources; Social Stigmatization; Labor Inequality; Violence and Crime; Lack of Political Representation, among others. Furthermore, class-based discrimination can intersect with other forms of discrimination, such as racism or sexism, exacerbating the difficulties faced by individuals belonging to multiple marginalized groups.

\subsection{Original prompt and examples} \label{subsec:prompt}
In this subsection, we present the original prompt and the provided examples for the few-shot scenario, both in Spanish. \textit{Instrucción} stands for Instruction, \textit{Ejemplos} for Example, \textit{Contexto}, \textit{Texto} and \textit{Salida} for Context, Text, and Output.\\

\textbf{Instrucción:}
"Determinar si el siguiente texto, correspondiente a un tweet y presentado con un contexto, contiene discurso de odio. Entendemos que hay discurso de odio si tiene declaraciones de carácter intenso y/o irracional de rechazo, enemistad o aborrecimiento contra un individuo o contra un grupo, siendo el individuo o el grupo objetivo de las expresiones por poseer una característica protegida. Las características protegidas que contemplamos son:
\begin{itemize}
    \item mujer: hace referencia a mujeres o al movimiento feminista.
    \item lgbti: hace referencia a gays, lesbianas, transexuales y otras identidades de género.
    \item racismo: hace referencia a inmigrantes, xenofobia, o a pueblos aborígenes.
    \item clase: hace referencia a personas de bajos recursos o motivos de clase.
\end{itemize}

Los tweets están escritos en español rioplatense, y dentro del contexto cultural de Argentina. Responder una o varias de las características separadas por coma, o bien "nada" si no hay discurso de odio. Pensar y justificar paso a paso la respuesta antes de responder."

\textbf{Ejemplos:}

\begin{quote}
    \textbf{contexto:} 'Wuhan celebra el fin de la cuarentena por el coronavirus con un mensaje para el resto del mundo: "Aprendan de nuestros errores"'

    \textbf{texto:} "Chinos culiados de mierda ojalá se mueran todos"

    \textbf{salida:} 'El texto insulta a los chinos ("culiados") y desea que se mueran, implicando que son los culpables del COVID-19. La respuesta final es "racismo".'\\

    \textbf{contexto:}"Al borde del llanto, Nati Jota pidió disculpas por sus repudiables tuits pero los justificó: 'Quería sumar seguidores'"

    \textbf{texto:} "Show o nada mi amorrr!! Sos una imbesil!"

    \textbf{salida:}  'El texto contiene un insulto pero no menciona nada relacionado a las características protegidas, pese a mencionar a una mujer. La respuesta final es "nada".'\\

    \textbf{contexto:}'Funcionario público y drag queen: la "doble vida" del político que representa a Salta en tacos, peluca y barba con brillos',

    \textbf{texto:}  "Que asco..!'",

    \textbf{salida:}'El texto expresa desagrado hacia la identidad de género de la persona. La respuesta final es "lgbti"  \\

    \textbf{contexto:} "Violento motín en la cárcel de Devoto: tomaron e incendiaron pabellones",

    \textbf{texto:} "Hermanito lo que son esos grones. Ahí adentro le robaron la billetera al coronavirus"

    \textbf{salida:} 'El texto utiliza la palabra "grones", que significa negros (se obtiene "grones" a partir de "negros" cambiando el orden de las sílabas). Negro o grone es una manera despectiva de refererirse a alguien por su color de piel. Además, da a entender que los presos son ladrones. La respuesta final es "racismo".' \\

    \textbf{context:}'Mayra Mendoza criticó el banderazo contra el Gobierno: "No tiene sentido y es antidemocrático"'

    \textbf{text:}  "Que negrita villera....ni con la plata q roba puede arreglarse"

    \textbf{output:} 'El texto alude al color de piel de Mayra Mendoza ("negrita") y a su condición social ("villera") de manera discriminatoria. La respuesta final es "racismo, clase".'\\

    \textbf{context:} "No da tregua: sin ropa y desafiante Flor Peña volvió a mostrar su lado más hot"

    \textbf{text:} "K lado hot tiene más varices k mi abuela. Tiene mas puesta de espalda k karagian trola"

    \textbf{output:}  'El texto alude al aspecto de Flor Peña (sus várices) y la desprecia por "trola", un insulto que significa "prostituta". La respuesta final es "mujer".'\\

    \textbf{context:} "Otro ataque de furia contra un colectivero: una mujer trans le pateó la unidad y le tiró un ladrillazo"

    \textbf{text:} "Un tipo operado. Con la fuerza de un hombre y no la de una mujer"

    \textbf{output:}  "salida": 'El texto asegura que la mujer trans es "un tipo operado", implicando que en los hechos es un hombre y negando su identidad de género. La respuesta final es "lgbti"' \\

    \textbf{context:}'Elisa Carrió denunció que el Gobierno usa la pandemia para "establecer un estado de sitio"'

    \textbf{text:} "Gorda psiquiátrica"

    \textbf{output:}  'El texto habla de su aspecto (gorda) y la acusa de tener problemas psiquiátricos. Sin embargo, no hace ninguna alusión a una característica protegida. La respuesta final es "nada".\\

    \textbf{context:} 'Loly Antoniale mostró su impresionante casa en Miami: "Soy la reina de mi castillo"'

    \textbf{text:} "No pudo enganchar al viejo famoso..se busco otro..y este le puso su castillo en Miami..."

    \textbf{output:} 'El texto alude a la vida amorosa de Loly Antoniale, sugiriendo que busca hombres por interés económico. Esta apreciación es sexista y despectiva. La respuesta final es "mujer".' \\

    \textbf{context:} "Les darán DNI provisorio a personas en situación vulnerable, para que puedan empezar a cobrar planes sociales"

    \textbf{text:} "Seguimos alimentando vagos"

    \textbf{output:}  'El texto se refiere a quienes cobran planes sociales como vagos a quienes hay que alimentar. Esto muestra un desprecio hacia las clases bajas que requieren de la asistencia estatal. La respuesta final es "clase".' \\

    \textbf{context:} "Los dos presos heridos de bala en el motín de Devoto tienen Covid-19 y uno quedó hemipléjico"

    \textbf{text:} "justicia divina!"

    \textbf{output:} "salida": "El comentario contiene una celebración frente a un ataque a presos. A pesar de ser sumamente ofensivo, no hace ninguna alusión a una característica protegida. La respuesta final es 'nada'." \\
\end{quote}